\documentclass[a4paper]{article}

\usepackage{spconf,amsmath,graphicx}
\usepackage{cite}
\usepackage{authblk}

\title{DATA AUGMENTATION FOR ROBUST KEYWORD SPOTTING UNDER PLAYBACK INTERFERENCE}
\makeatletter
\def\@name{ \emph{Anirudh Raju$^{1}$, Sankaran Panchapagesan$^{2*}$,\thanks{*Work conducted while the author was at Amazon.com}, Xing Liu$^{3}$, Arindam Mandal$^{1}$, Nikko Strom$^{1}$}}
\makeatother
\address{${}^{1}$Amazon Alexa, USA \\  ${}^{2}$Google Inc., USA \\ $^3$Purdue University, USA \\ \{ranirudh,arindamm,nikko\}@amazon.com, panchi@google.com}

\begin{document}

\maketitle

\begin{abstract}
Accurate on-device keyword spotting (KWS) with low false accept and false reject rate is crucial to customer experience for far-field voice control of conversational agents. It is particularly challenging to maintain low false reject rate in real world conditions where there is (a) ambient noise from external sources such as TV, household appliances, or other speech that is not directed at the device (b) imperfect cancellation of the audio playback from the device, resulting in residual echo, after being processed by the Acoustic Echo Cancellation (AEC) system. In this paper, we propose a data augmentation strategy to improve keyword spotting performance under these challenging conditions. The training set audio is artificially corrupted by mixing in music and TV/movie audio, at different signal to interference ratios. Our results show that we get around 30-45\% relative reduction in false reject rates, at a range of false alarm rates, under audio playback from such devices.
\end{abstract}

\section{Introduction}
Virtual assistants typically deploy a small-footprint keyword spotting system on-device which listens for a specific keyword such as  "Alexa" \cite{sane_talk, kumatani2012microphone, kumatani2017direct, Guo2018, sun2016max}. On detecting the keyword, the device can stream up the data to the cloud and proceed to recognize and interpret voice commands. Since each interaction with the device is prefaced with the keyword, accurate detection is very important for a great customer experience. In an ideal scenario, the device would wake up whenever addressed by the keyword (i.e. low false reject rate or FRR), and never wake up unless addressed to (low false alarm rate or FAR). In addition, the KWS system should be able to detect words spoken in natural conversational speech in far-field settings, irrespective of where the device is placed in the room or the actual dimensions of the room. The system should be robust to external sound sources in the room such as TV or music playing in the background, noise from refrigerators, AC vents, dishwashers, etc, and other speech in the room not directed to the device.

The user may wish to interrupt and wake up the device when it is playing (a) music, radio, etc (b) Text-to-Speech (TTS) Output. This condition is referred to as barge-in under playback. An Acoustic Echo Cancellation (AEC) algorithm \cite{hansler2004acoustic} is typically run on the device to cancel the playback signal from the speech + playback signal that is captured by the microphones. It's used as a front-end signal processing technique to preprocess the input audio signal before feeding into the keyword spotting engine. AEC assumes that the acoustic transfer function from the loudspeaker to the microphone is a deterministic linear model \cite{paleologu2013study}. When the playback speaker or front end signal processing introduces non linear artifacts, it may diverge the filter estimate. Moreover, in certain cases such as audio playback from an external bluetooth speaker, it is difficult to obtain time-aligned reference playback information. These limitations reduce the beneficial impact of AEC in terms of Echo Return Loss Enhancement (ERLE), and results in an audio stream that contains some residual playback information. Hence, it is important that the keyword spotting system is robust to audio signals where the speech signal is mixed with the playback signal.

Noise robustness techniques for speech recognition \cite{virtanen2012techniques} and keyword spotting systems commonly fall into the categories of (a) signal or feature enhancement techniques that aim to remove the noise from the input features \cite{loizou2013speech, xu2014experimental} (b) model adaptation techniques that modify the model parameters to be more representative of the input speech \cite{kalinli2009noise} and (c) feature adaptation techniques such as fMMLR \cite{gales1998maximum}. In addition to these techniques, data augmentation has been used to increase quantity of training data, avoid overfitting and increase robustness of models \cite{seltzer2012acoustic, hsiao2015robust, hannun2014deep, ko2015audio}.

In this paper, we are primarily concerned with improving the robustness of the small-footprint KWS system to noise from residual playback information due to imperfect AEC. The main contribution of this paper is to pose this a noise robustness problem as opposed to an AEC signal enhancement problem, by using a data augmentation strategy under an additive noise model. In addition, we see improved robustness to external sound sources that may have similar characteristics to device audio playback. This adds no additional complexity to the small-footprint KWS system during runtime, keeping in mind that our system runs on device, which places constraints on both runtime memory usage and CPU cycles.

The rest of the paper is organized as follows. In Section 2, the Alexa KWS system is reviewed and the proposed data augmentation method is described. This is followed by the experimental setup in Section 3. The results along with their interpretations are presented in Section 4, and a summary is provided in the final section.

\section{The Keyword Spotting System} 
\label{section:kws_system}
\begin{figure}[h]
\vskip 0.2in
\begin{center}
\centerline{\includegraphics[width=\columnwidth]{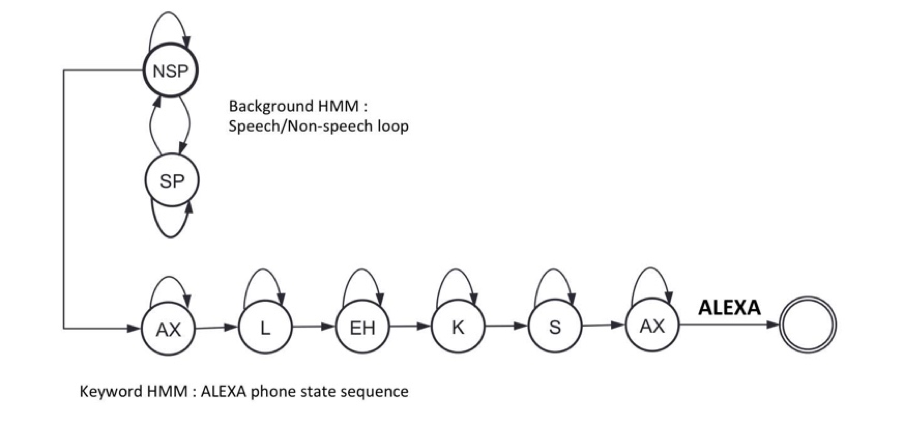}}
\caption{HMM-based Keyword Spotter}
\label{fig:system}
\end{center}
\vskip -0.2in
\end{figure}

We are primarly interested in building a small-footprint KWS system for a single predefined keyword. A common approach is to use Hidden Markov Models (HMMs) to model both the keyword speech and the background non-keyword audio \cite{wilpon1990automatic, rose1990hidden, wilpon1991improvements}. The background model is also known as a garbage or filler model in the literature. It can be modeled using a simple HMM with loops over speech/non-speech phones \cite{wilpon1990automatic} or in complicated cases, with loops over all phones in the phone set or words \cite{weintraub1993improved}. The foreground model is an HMM with the phone state sequence corresponding to the keyword of interest. In Figure \ref{fig:system}, we show an example Finite State Transducer (FST) architecture of an HMM-based keyword spotter for the keyword "Alexa". The background HMM is modeled using a single HMM state for non-keyword speech and another HMM state for non-speech. There are six phones in the "Alexa" keyword, each of which is represented in the foreground HMM by a single state HMM for the sake of simplicity. In practice, it is beneficial to model these using three or five state HMMs. 

In our system, we use a Deep Neural Network (DNN) based acoustic model \cite{hinton2012deep} in combination with the HMM decoder. The DNN provides acoustic scores over each of the foreground and background HMM states, for each acoustic frame input. A keyword is hypothesized when a final state of the KW FST is reached.

\section{Data Augmentation Strategy}
Data augmentation aims to artificially corrupt the training data at different signal-to-noise ratios so that the training data has a generalized representation of the data to be processed in the actual user case. It has been observed that the performance of a DNN acoustic model depends on how well the training data matches the testing data. The representative characteristics that are seen in real test data, such as different background noise conditions, speaker variability, channel variability should be included in the training data so that the DNN can learn to be robust to these variations. 

In general, by training the DNN on multi-condition data, we enable it to learn features that are more invariant to these different types of noise, with respect to the classification accuracy measured through the cross-entropy objective function. The lower layers of the network can be viewed at as a non-linear feature extractor, which attempt to discriminatively learn a feature representation that is invariant to the different acoustic conditions in the training data.
 
A DNN that is trained on data with a high signal-to-noise (SNR) ratio overall, would result in degraded performance when tested on noisy data. Hence, when dealing with speech recognition task in a noisy environment, it is desired that the DNN model is trained with training data that is corrupted in a similar fashion.

\subsection{Corrupted Data Preparation}

The training data set for the keyword DNN acoustic model consists of relatively clean far-field microphone data. We artificially corrupt this data by adding reverberated music or TV/movie audio to each utterance at a certain speech-to-interference (SIR) ratio. The movie audio is obtained from an Amazon internal dataset, which contains TV shows and movies. It may include title/background music, dialogue and other sound effects. The music audio is obtained from a list of popular songs. This aims to simulate two very realistic use cases where (a) audio (maybe music, radio, podcasts, etc) is being played from the device, and is not perfectly cancelled by the echo cancellation algorithm (b) music/video is played from a background sound source while people are attempting to talk to the device. 

The corrupted data that is used for DNN training can be expressed as

\begin{equation}
\label{eqn:alpha_SIR}
y_{corrupted}(t) = x_{utterance}(t) + \alpha  \cdot x_{interference}(t) 
\end{equation}
where $x_{utterance}(t)$ is the original uncorrupted utterance audio,  $x_{interference}(t)$ is the interference signal and $\alpha$ is the scaling factor applied to the interference

For each utterance, the interference signal $x_{interference}(t)$, is obtained by reverberating the movie/music data. This is done through a convolution with a room impulse response (RIR), as shown in Eqn \ref{eqn:convolution}. The RIRs used in our experiment are collected by playing and recording chirp signals in a room. This artificial reverberation helps us simulate the audio captured by the far-field microphones. In order to simulate real world usage conditions, we obtain RIRs from several different acoustic environments.

\begin{equation}
\label{eqn:convolution}
x_{interference}(t)=x_{music}(t)*g_{RIR}(t)
\end{equation}
where $g_{RIR}$ is the room impulse response, $x_{music}$ is the movie/music audio that is used to approximate the (a) playback or (b) external noise source

To simulate the environment where the speech is mixed with noise at different levels, we scale the reverberated audio according to a random Signal-to-Interference ratio (SIR) in a predefined range before adding it to the utterance. The SIR over the utterance is defined by

\begin{equation}
\label{eqn:SIR}
SIR (dB) = 20 log \frac{\sqrt{\sum_{i=0}^{N}s^{2}[i]}}{\alpha\sqrt{\sum_{i=0}^{N}n^{2}[i]}}
\end{equation}

where $s[i]$ are speech samples from one utterance, $n[i]$ are interference segments that are randomly selected from the reverberated movie/music, $N$ is the number of samples in the utterance, and $\alpha$ is the scale applied to the interference. From the above Eqn \ref{eqn:SIR}, $\alpha$ can be computed:

\begin{equation}
\label{eqn:alpha_SIR}
\alpha =  \frac{\sqrt{\sum_{i=0}^{N}s^{2}[i]}}{\sqrt{\sum_{i=0}^{N}n^{2}[i]}} 10^{-\frac{SIR (dB)}{20}}
\end{equation}

The targets for DNN training are obtained by running forced alignments of the ground truth sequence of HMM states, on the uncorrupted utterance data. 

Note that the data augmentation can be extended based on the use case of the device, to any other type of interference: background speech, noise from household appliances or air conditioning, car noise, gaussian noise, white noise, etc.

\begin{figure}[ht]
\begin{center}
\centerline{\includegraphics[width=\columnwidth]{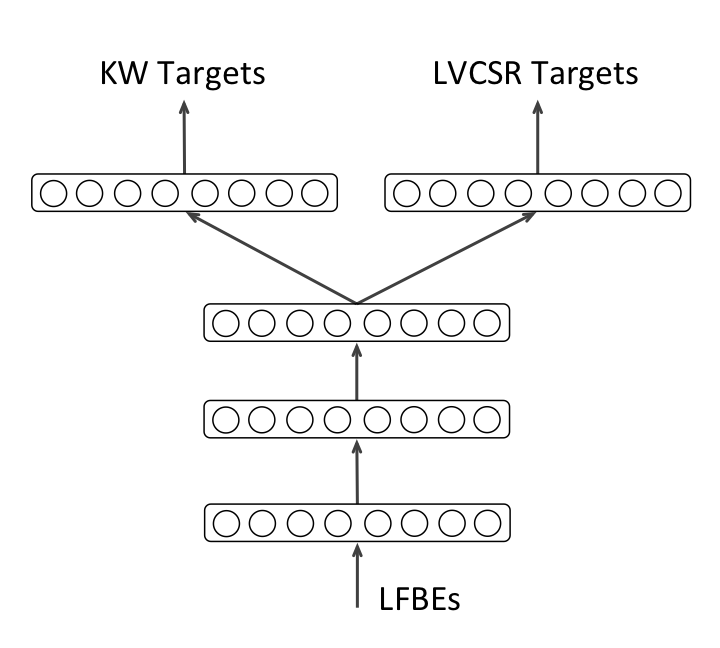}}
\caption{DNN Architecture}
\label{fig:multitask}
\end{center}
\vskip -0.2in
\end{figure} 

\begin{figure}[h]
\begin{center}
\centerline{\includegraphics[width=\columnwidth]{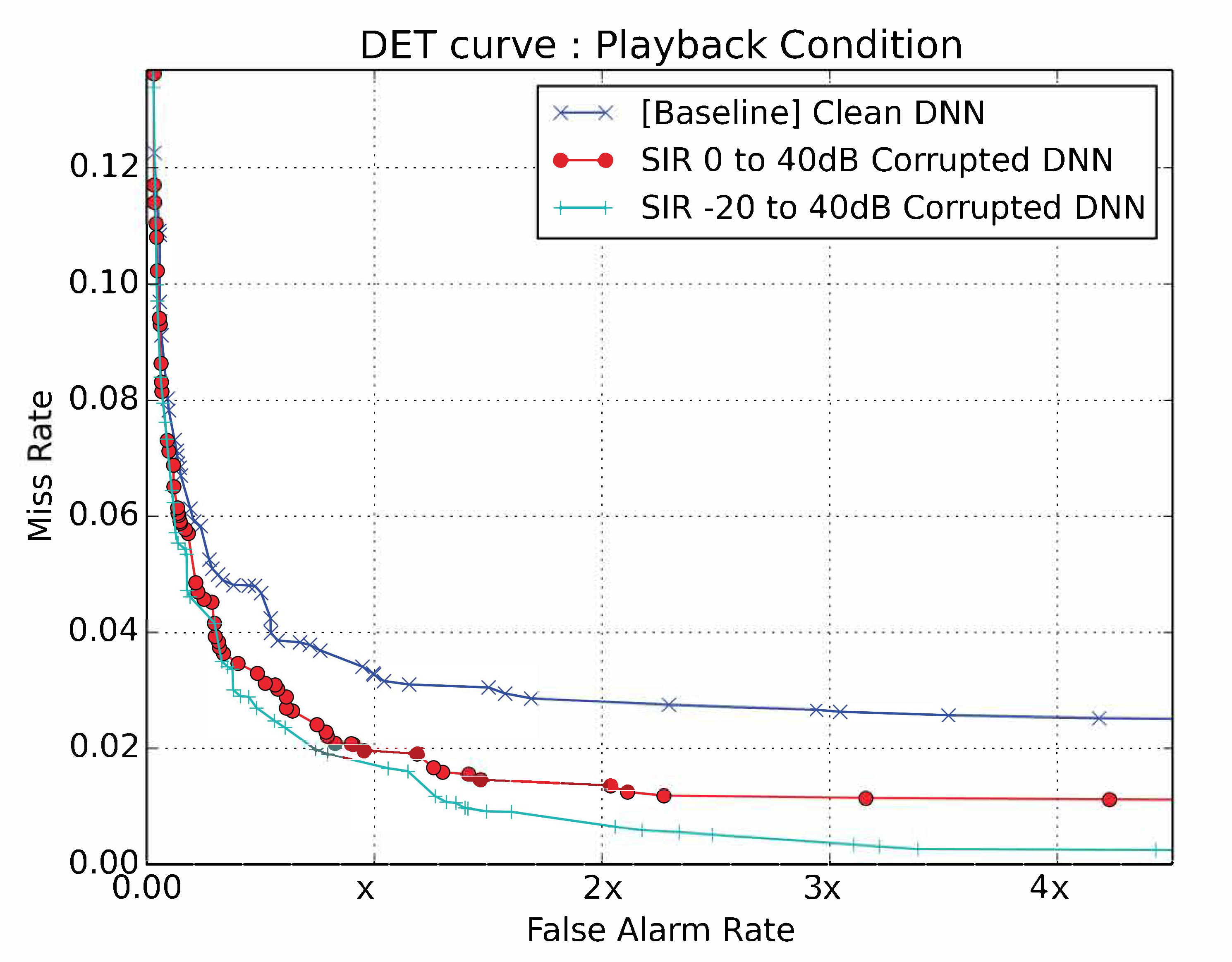}}
\caption{Comparison between DNNs trained on clean and noisy data, corrupted with two different SIR ranges, on a test set with playback} 
\label{fig:int-v3_309}
\end{center}
\vskip -0.2in
\end{figure} 

\begin{figure}[!h]
\begin{center}
\centerline{\includegraphics[width=\columnwidth]{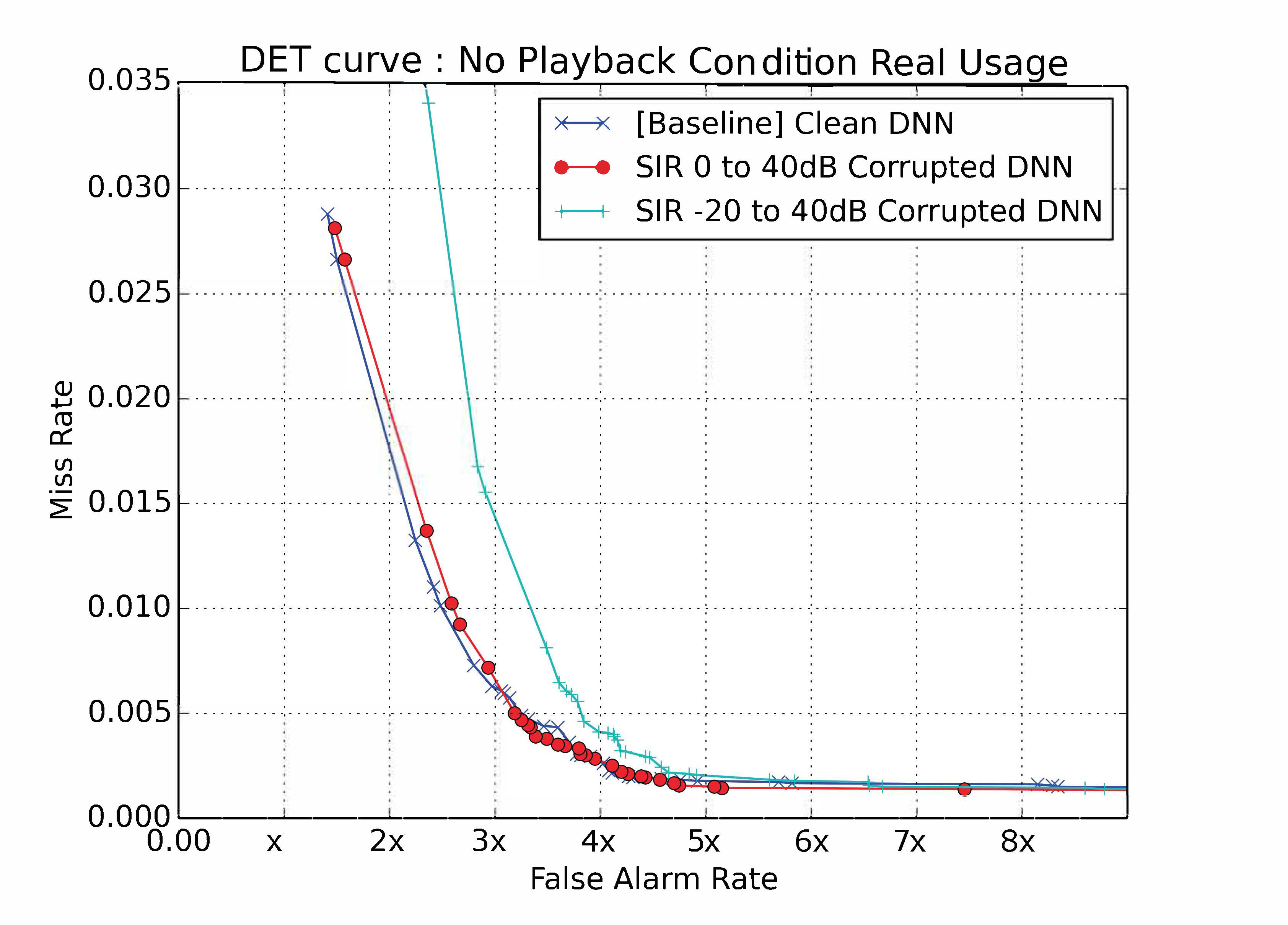}}
\caption{Comparison between DNNs trained on clean and noisy data at different SIR levels on a test set without playback but captured through far-field microphones} 
\label{fig:lv_309}
\end{center}
\vskip -0.2in
\setlength{\belowcaptionskip}{-10pt}
\end{figure}

\section{Experimental Setup}

The experimental results shown here are based on the DNN-HMM hybrid keyword spotting system described in Section \ref{section:kws_system}. The structure of the DNN is shown in Fig \ref{fig:multitask}. The DNN is trained to predict the large vocabulary continuous speech recognition (LVCSR) task, in addition to the keyword task, since this is known to improve the accuracy of the 1st stage detection \cite{panchi-multitask}. The deep learning trainer described in \cite{strom2015scalable} is used to train the DNN on GPUs. A weighted cross-entropy objective function is used where the loss stream corresponding to the speech recognition targets are weighted at 0.1 and the loss stream corresponding to the keyword targets are weighted at 0.9.

\subsection{Training and Test Datasets}
The models for all our experiments are trained with ~3000 hours of far-field acoustic data. The original 3000hr dataset, which contains relatively clean data, is corrupted with (a) Movie and (b) Music data to produce two more training datasets. The corrupted training datasets are both 3000hr each. The targets for the DNN training are obtained by aligning the clean data with an ASR model, and mapping the corresponding senones to the keyword phones. 
We evaluate on the following test sets

\begin{enumerate}
	\setlength\itemsep{0em}
	\item Playback Condition
	\item No AEC Condition (With Playback) \label{item:no-aec}
	\item No Playback Condition Real Usage
	\item Mixed Conditions Real Usage
\end{enumerate}

All of the above test sets are after processing with the AEC front-end, except for test set \ref{item:no-aec}. These test sets each contain several hundreds to thousands of keywords and potential false alarms. The values of the False Alarm Rate (FAR) obtained through these test sets is an overestimate of the in-field FAR because we only collect the utterances for which the device wakes up. However, relative improvements in the first stage FAR on these test sets, result in improved end to end performance on the field.

\section{Results and Discussion}

The performance of the keyword system is presented here in the form of DET curves which are plots of the false reject rate vs the false alarm rate. The different operating points are obtained by tuning over the HMM parameters, and selecting the best possible false reject rate at a given false alarm rate. We also present Area Under the Curve (AUC) numbers to help us quantify the performance improvement of these models. Note that since we present AUC numbers for DET curves instead of ROC curves, lower AUC numbers correspond to better performance.

\begin{figure}[h]
\begin{center}
\centerline{\includegraphics[width=\columnwidth]{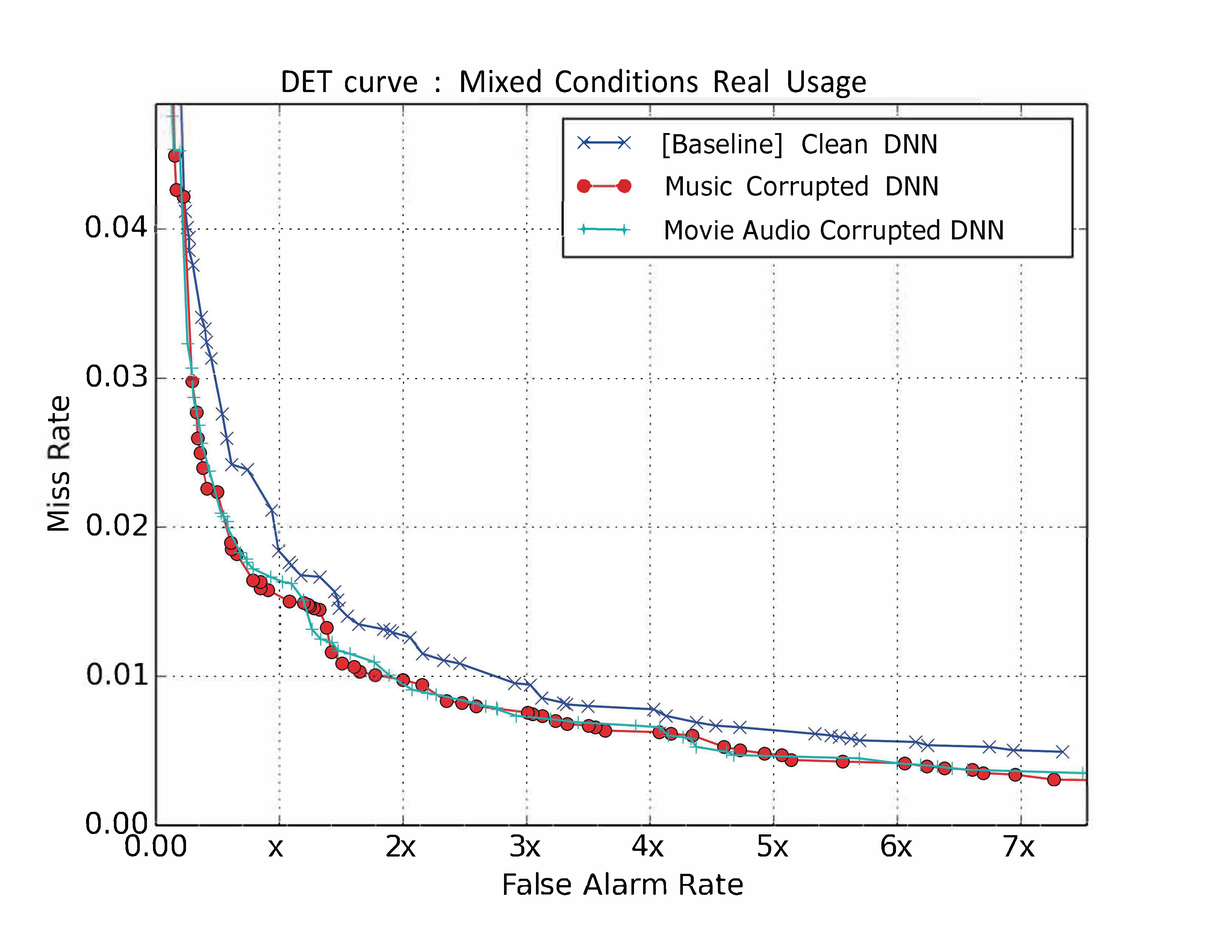}}
\caption{Comparison between DNNs on a test set that contains utterances with and without playback. It comes from real usage of the device in people's homes, so the fraction of playback to no-playback utterances is what we see on the field} 
\label{fig:b_nm}
\end{center}
\vskip -0.2in
\end{figure} 

\begin{figure}[h]
\begin{center}
\centerline{\includegraphics[width=\columnwidth]{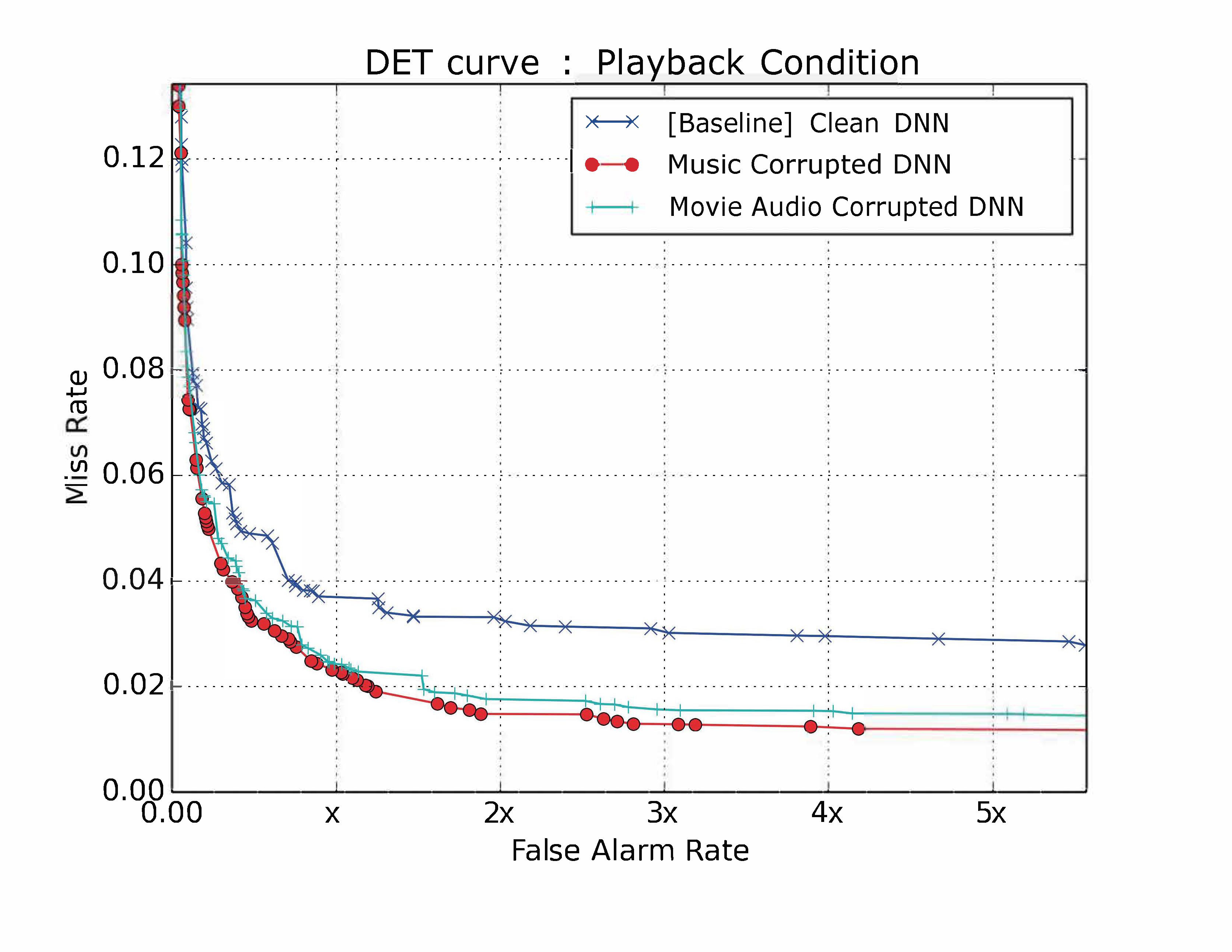}}
\caption{Comparison between DNNs on a test set with playback} 
\label{fig:int-v3}
\end{center}
\vskip -0.2in
\end{figure}

\subsection{Varying levels of interference}

The corrupted data is prepared by mixing each utterance in the training dataset with a randomly cropped reverberated movie/music segment, at a randomly chosen Signal-to-Interference (SIR) in the predefined range. SIR ranges of [0, 40] dB and [-20, 40] dB
have been explored. A uniform distribution has been used for sampling the SIRs. All of these models are trained on a subset of the full 3000h corpus (1000hrs).

Fig \ref{fig:int-v3_309} and Fig \ref{fig:lv_309} show the first stage keyword performance, on test sets with and without playback respectively. Here, we see that data corruption with both SIR ranges of [-20, 40] dB and [0, 40] dB result in an improvement on the playback test set. The more aggressive data corruption with [-20, 40] dB results in a much larger improvement on the playback test set but a degradation on the no playback test set. Since we aim to improve keyword performance under playback/interference without significant degradation under other conditions, we select the [0, 40] dB SIR range for the experiments going forward.

\subsection{Data Corruption with Movie vs Music data}
\begin{table}[h]
\caption{Area Under the Curve (AUC) comparison of DET plots from Figure \ref{fig:int-v3}. (Range of FARs used to compute AUC was 0.001 to 0.05). Note that since these are computed on DET curves, a lower value of AUC is better}
\label{auc-table}
\vspace{-2mm}

\begin{center}
\begin{small}
\begin{sc}
\begin{tabular}{|p{4cm}|p{1cm}|p{2cm}|}
\hline
Model & AUC & \% Reduction \\
\hline
[Baseline] Clean DNN   & 0.170 & 0.0 \%\\
Movie Corrupted  & 0.102 &  40.0 \%\\
Music Corrupted  & 0.089 & 47.6 \% \\
\hline
\end{tabular}
\end{sc}
\end{small}
\end{center}
\end{table}

Next we compare the effects of corrupting the training data with movie vs music data. From Figure  \ref{fig:b_nm}, we see that  models trained on data corrupted with both movie and music result in similar accuracy improvements under mixed real usage conditions. However, from Figure \ref{fig:int-v3}, we see that corruption with music gives slightly larger accuracy improvements on the playback test set. These improvements are better quantified in table \ref{auc-table}. We see that we get a 47.65\% reduction in AUC from using the music corrupted model, over the baseline clean model.

\subsection{Results with no AEC}

\begin{figure}[ht]
\begin{center}
\centerline{\includegraphics[width=\columnwidth]{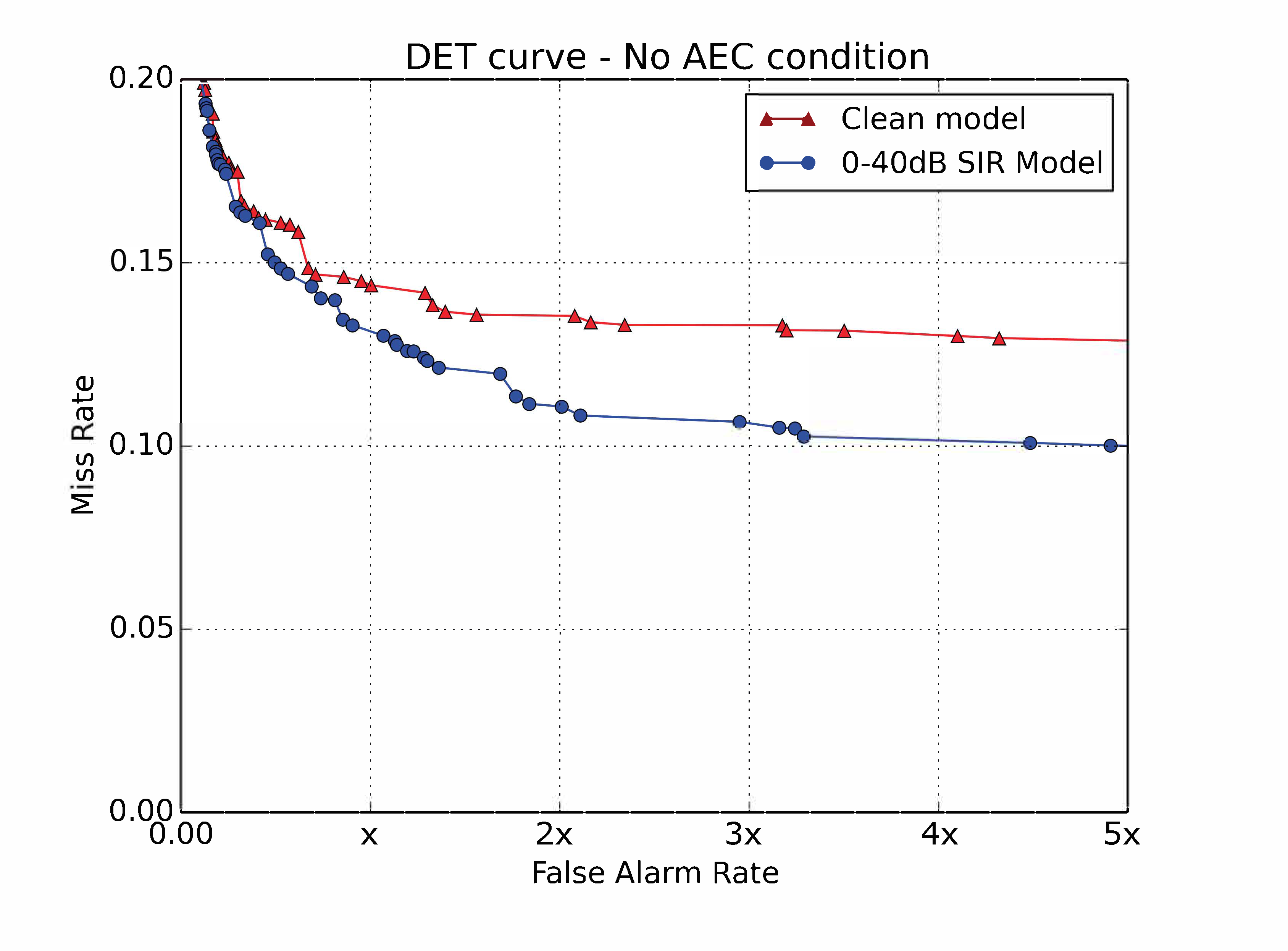}}
\caption{Comparison between DNNs on a test set that has not been processed with the Acoustic Echo Cancellation (AEC) algorithm} 
\label{fig:no-aec}
\end{center}
\vskip -0.2in
\end{figure}

In certain situations, the AEC front-end signal processing could be rendered ineffective due to channel conditions, or ambient playback that we cannot attenuate using AEC (such as TV playback). In Fig \ref{fig:no-aec}, we study the effect of data augmentation on a test set that hasn't been processed by the AEC algorithm. We see that data augmentation is beneficial even under these conditions.

\section{Summary}

We described the challenges in maintaining a small-footprint KWS system with low false rejection rate under playback conditions and in the presence of interfering sound sources. In particular, we improved the robustness of this system to noise from residual playback due to imperfect Acoustic Echo Cancellation (AEC). We posed this as noise robustness problem as opposed to an AEC signal enhancement problem, and proposed a data augmentation strategy under an additive noise model. This enabled us to improve the performance of our system while adding no additional complexity in terms of memory or CPU usage during runtime.

Our results show that we are able to significantly improve the performance of the KWS system under playback conditions irrespective of whether the data has been processed through an AEC system. In particular, we obtained a 47\% relative reduction in AUC using data augmentation with music data and 40\% relative reduction in AUC with movie data corruption. The lower false rejection rate under real usage conditions leads to an improved customer experience with the far-field conversational agent.

In addition, we have extended these techniques to newer neural modeling architectures based on raw audio input, for keyword spotting, where they show significant accuracy improvements \cite{kumatani2017direct}.

\bibliographystyle{IEEEtran}
\bibliography{mybib}

\end{document}